# A Masked language model for multi-source EHR trajectories contextual representation learning

Ali AMIRAHMADI[a,1], Mattias OHLSSON[a,b], Kobra ETMINANI[a], Olle MELANDER[c], and Jonas BJÖRK[d]

[a] *Center for Applied Intelligent Systems Research, Halmstad University*
[b] *Centre for Environmental and Climate Science, Lund University*
[c] *Division of Occupational and Environmental Medicine, Lund University*
[d] *Department of Clinical Sciences, Lund University*

ORCiD ID: Ali Amirahmadi https://orcid.org/0000-0002-1999-8435

**Abstract.** Using electronic health records data and machine learning to guide future decisions needs to address challenges, including 1) long/short-term dependencies and 2) interactions between diseases and interventions. Bidirectional transformers have effectively addressed the first challenge. Here we tackled the latter challenge by masking one source (e.g., ICD10 codes) and training the transformer to predict it using other sources (e.g., ATC codes).

## 1. Introduction

Electronic health records (EHRs) together with advanced machine learning, have provided opportunities for the next generation of medical decision support systems. The longitudinal collection of EHR data (aka. patient trajectories) has successfully been analyzed using recent developments in natural language processing methods. Pre-trained language representations with self-supervised methods have found their counterpart for EHR data, such as EHR2Vec and EHR contextual learning(1). Language models such as Bert and their EHR peers like Behrt(2) and MedBert(3),because of their capability to handle time dependencies have outperformed the state-of-the-art methods for predicting medical outcomes. In this work we addressed the challenge of interactions between different data sources using new masking methods for Masked language models learning.

## 2. Material and methods

This paper used the history of diagnoses (1.5M ICD10 codes), medications (6M ATC codes), and two questionnaires (baseline and five year follow-up with 1895 questions in total) of approximately 30,000 persons from the Malmö Diet and Cancer Cohort(4).

---

[1] Corresponding Author: Ali Amirahmadi, ali.amirahmadi@hh.se

We devised a two-step masking process to learn the effective representation of the multi-source EHR data and address mentioned challenges. Firstly, to handle the long/short-term dependencies, we randomly mask some proportion of the patient trajectories and train our network to predict the masked part via the unmasked ones. Then, to model the interactions between life habits, interventions, and diseases, we mask one of the sources for a specific period and train the network to predict it using other sources during that period.

Then to learn the representation, we train a transformer encoder to predict the masked part of the input sequence data (**Figure 1**). In the last step, we add a classifier layer on top of the trained network and fine-tune it for downstream tasks.

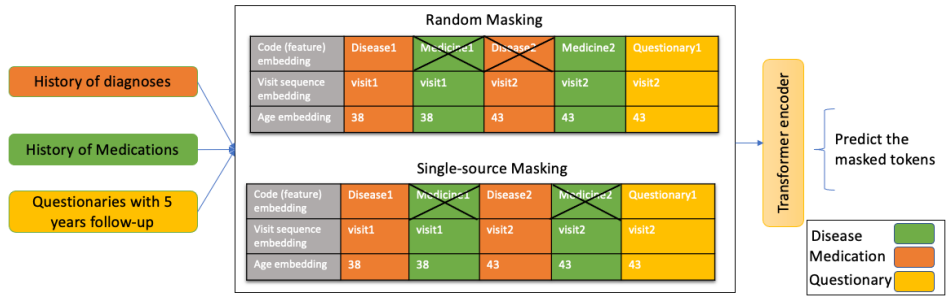

**Figure 1.** Learning the effective representation of multi-source patient trajectories by a two-step Masking MLM and a multi-head transformer encoder

## 3. Results, Discussion, and Conclusion

To evaluate the built representation learning model, we fine-tuned it to predict the Heart-failure ICD10 codes in the next visit. Table 1 show the promising ability of the two-step MLM.

Here we introduced a two-step masking process for patient trajectories representation learning. Developing an effective ML model strongly depends on efficient feature extraction from all available sources. We expect the developed model to effectively address the time dependencies and the interactions between disease, interventions, and other sources simultaneously. Incorporating data sources with other modalities like medical images still needs more investigation.

Table 1. The average AUC values and standard deviations of 5-fold CV (in parentheses) for HF in the next visit on test data

| Used data / method | Logistic regression | Random Forest | MLP | Bi-GRU | Simple-MLM + Bi-GRU | Design-masking Bert |
|---|---|---|---|---|---|---|
| diagnoses | 0.593 (0.005) | 0.526 (0.001) | 0.685 (0.005) | 0.666 (0.004) | 0.765 (0.012) | - |
| diagnoses and medications | 0.644 (0.004) | 0.564 (0.004) | 0.752 (0.005) | 0.783 (0.012) | 0.909 (0.002) | 0.919 (0.006) |